\documentclass{llncs}
\usepackage{makeidx}  

\usepackage{cite}
\usepackage[tight,footnotesize]{subfigure}
\usepackage{algorithm}
\usepackage{algorithmic}
\usepackage{psfrag}
\usepackage{amsmath}
\usepackage{mathptmx}       
\usepackage{helvet}         
\usepackage{courier}        
\usepackage{type1cm}        
\usepackage{makeidx}         
\usepackage{graphicx}        
\usepackage{multicol}        
\usepackage{multirow}        
\usepackage[bottom]{footmisc}
\usepackage{color}

\begin{document}
%

\mainmatter              
\title{Deep Learning in Multi-Layer Architectures of Dense Nuclei}
\titlerunning{Multi-Layers with Dense Recurrent Nuclei}  
%
\author{Yonghua Yin and Erol Gelenbe}
\authorrunning{Yonghua Yin and Erol Gelenbe} 
%
%
\institute{Intelligent Systems and Networks Group,
Electrical \& Electronic Engineering Department,
Imperial College, London SW7 2AZ, UK\\
\email{y.yin14@imperial.ac.uk, e.gelenbe@imperial.ac.uk}}

\maketitle              

\begin{abstract}
We assume that, within the dense clusters of neurons that can be found in nuclei, cells may interconnect via soma-to-soma interactions, in addition to conventional synaptic connections. We illustrate this idea with a multi-layer architecture (MLA) composed of multiple clusters of recurrent sub-networks of spiking Random Neural Networks (RNN) with dense soma-to-soma interactions, and use this RNN-MLA architecture for deep learning. The inputs to the clusters are first normalised by adjusting the external arrival rates of spikes to each cluster. Then we apply this architecture to learning from multi-channel datasets. Numerical results based on both images and sensor based data, show the value of this novel architecture for deep learning.
%
\end{abstract}
\section{Introduction}  \label{sec.rnnmodel}

Deep learning has achieved great success \cite{hinton2006reducing,lecun2015deep,Deep1,Deep2}, through multilayer architectures that
extract high-level representations from raw data. However, as the number of neurons that are used increases, so does the computational complexity and memory requirements of the
algorithms that are used.
Thus in \cite{gelenbedeep2016}, we constructed network models for deep learning that exploit the asymptotic properties of very large clusters of cells by reducing them to
simplified transfer functions based on spiking Random Neural Networks (RNN) with multilayer architectures (MLA) and extreme learning machines (ELM) \cite{elm,mlelm}. The resulting
RNN-MLA architecture provided highly effective classification rates on real data sets with a reduced computational complexity as compared to other deep learning techniques.

In this paper we pursue the idea that the human brain contains important areas composed of dense clusters of cells, such as the basal ganglia and various nuclei. These clusters may be composed of similar or identical cells, or of varieties of cells. Because of their density, we suggest that such clusters may allow for a substantial amount of direct communication between stomata, in addition to the commonly exploited signalling that takes place through dendrites and synapses.

Thus we consider a network composed of a multi-layer structure (MLA)
where each layer is composed of a finite number of dense nuclei. Each nucleus is modelled as a recurrent spiking Random Neural Network \cite{RNN89}. Each neuron in each nucleus has a statistically identical interconnection structure
with the other cells in the same nucleus. This statistical regularity allows for a great individual variability among neurons both with regard to spiking times and the interconnection patterns. Within each nucleus the cells communicate with each other \cite{Timotheou} in a recurrent fully connected recurrent structure that can use both synapses and direct soma-to-soma interactions.

\begin{figure}[t]
\centering
\includegraphics[width=4in]{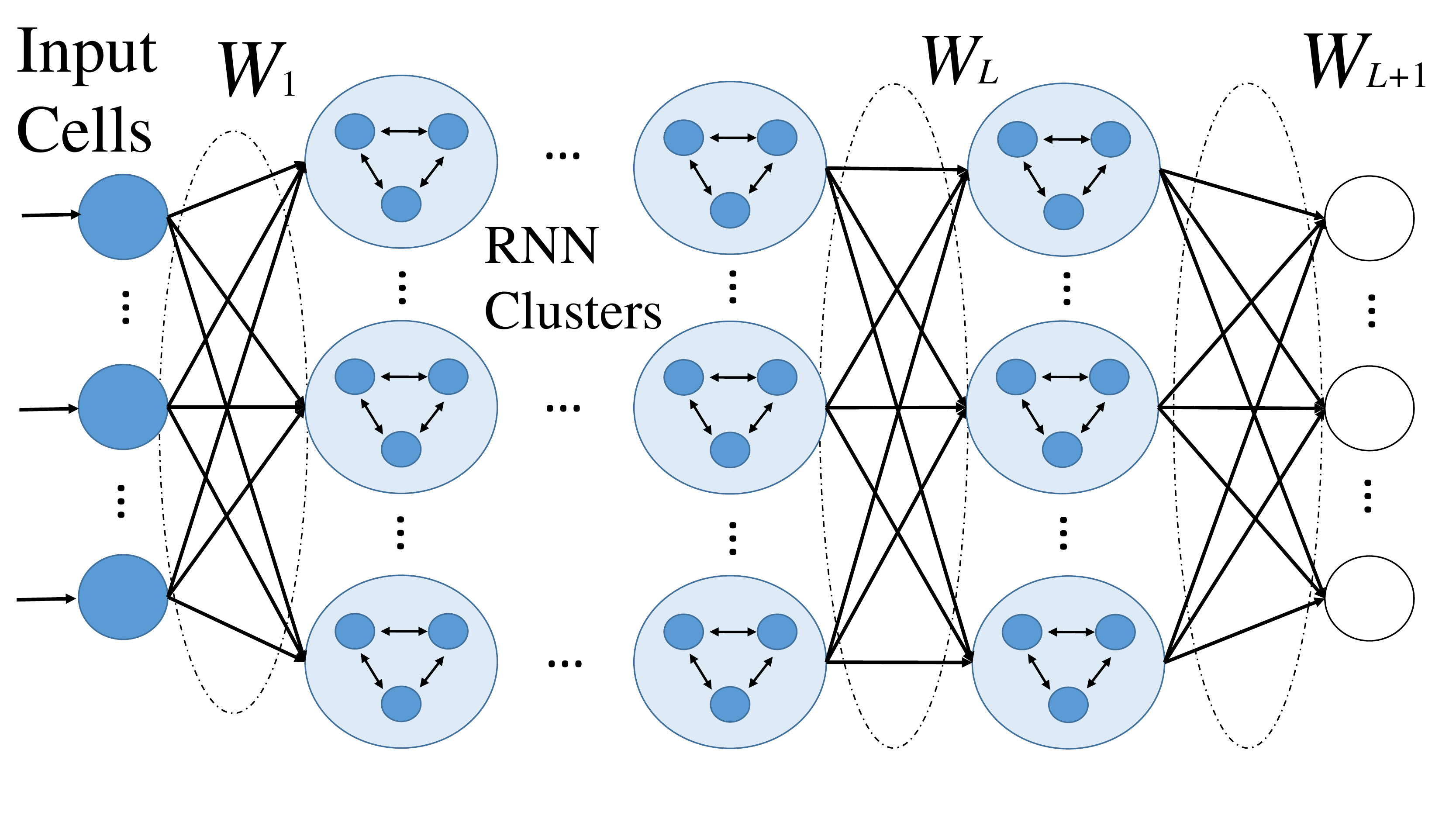}
\caption{Schematic representation of the RNN-MLA.}
\label{fig.RNN_MLA}
\end{figure}

On the other hand, the communication structure between different layers of nuclei is a conventional multi-layer feedforward structure, where
the nuclei in the first layer receive excitation signals from external sources, while each cell in each nucleus
has an inhibitory projection to the next higher layer. This RNN-MLA architecture is shown schematically in
Figure \ref{fig.RNN_MLA}.

\section{The Mathematical Model}

Small groups of spiking neurons can be represented conveniently using differential equation models. However, when large ensembles of hundreds or thousands of cells are represented,
as in the nuclei that we consider, probability models can be more convenient and tractable \cite{gelenbe1990stability,timotheou2010random}.
Thus in this paper we use the
Random Neural Network (RNN), which is a stochastic and recurrent model \cite{gelenbe1999random} that mimics a very large
``integrate and fire'' system \cite{Lapicque}, with cells that are represented by a discrete cell state. The emission of spikes occurs from a neuron when its discrete state drops by $1$ without the external arrival of an inhibitory spike.

In the RNN-MLA architecture, nuclei in the first (input) layer are made up of cells that receives excitatory spike trains from external sources, resulting in a linear cell activation $q(x)=x$. The successive $L$ layers are hidden layers composed of M$(n)$ clusters that receive inhibitory spike trains from cells in the previous layer, with a resultant activation function $q(x)=\zeta(x)$.

\subsection{Nuclei with Inhibitory Cells}

If a nucleus is composed of statistically identical inhibitory cells, then
using the RNN equations \cite{RNN89}, the excitation (activation) probability $q$ of any of
the statistically identical cells as a function of the external inputs $[\lambda^+,\lambda^-]$  becomes:
\begin{eqnarray}\label{inhibit}
q &=& \frac{\lambda^+}{r + \lambda^- + qr},\\\nonumber
&&or~~q^2 r + q(r +\lambda^-) - \lambda^+ = 0,\\\nonumber
&&yielding:~~q= \frac{\sqrt{(r+\lambda^-)^2 + 4r \lambda^+} - (r+\lambda^-)}{2r},
\end{eqnarray}
where $r$ is the firing rate of each cell, $\lambda^+$ and $\lambda^-$ are Poisson arrival rates of external and excitatory inhibitory spike trains to each cell.

\subsection{Activation only through Some-to-Soma Interactions} \label{soma}

Soma-to-Soma interactions occur when a cell in a cluster, say $C_1$ fires, and provokes the simultaneous or quasi-synchronous firing of other cells $C_2~...~,C_m$ which are also excited, and possibly leading to the excitation
of some other cell $C_{m+1}$ which in turn may fire later. As a result, the excitation level of cells $C_1,~...~,C_m$ drops, while the excitation level of cell $C_{m+1}$ rises.

Clearly, there may be many such patterns of communication, and the simplest occurs as follows in an $n\geq 3$-cell network.
Let us fix some cell,
say $C_1$,
and consider any other cell $C_2$, selected with probability $\frac{1}{n-1}$; if it is excited it will fire at rate $r$
and cause some other cell other than $C_1$, say $C_3$ now selected with probability $\frac{1}{n-2}$, to fire and together they will excite
$C_1$: this leads to the terms in the numerator of \ref{soma-fire}. The terms in the denominator of \ref{soma-fire}
result from the case where any excited cell such as $C_2$ fires at rate $r$ and triggers the firing of $C_1$.  In the RNN formalism, this leads to the
following representation of the excitation probability $q$ of cell $C_1$, assuming that all cells are homogenous and statistically identical:
\begin{eqnarray} \label{soma-fire}
&&q = \frac{\lambda^+ + (n-1)\frac{1}{n-1}q.r.(n-2)\frac{1}{n-2}q\frac{1}{n-1}}{r+\lambda^-+(n-1)\frac{1}{n-1}q.r\frac{1}{n-1}},\\\nonumber
&&yielding:~~~q[r+\lambda^-]=\lambda^+,~or~~q=\frac{\lambda^+}{r+\lambda^-}.
\end{eqnarray}
Thus this simple case is mathematically equivalent to a nucleus where neurons do not interact internally with each other at all.

\subsubsection{Random Selection of Soma-to-Soma Interactions}

Let us now consider a nucleus whose cells receive an inhibitory input from some external cell $u$ of the form $q_uw^-_u$where $q_u$ is the state of the external cell, and $w^-_u$ is the corresponding inhibiry weight.

In this case, the interconnect pattern for soma-to-soma interactions within a nucleus of $n$ cells proceeds as follows.
When a given cell fires, and then provokes repeated firing with probability $p$ among the
the subset of cells of size $n-1$ which contains the cell that first fired, and terminates with probability
$(1-p)$ by exciting the $n$-th cell which was not among the $n-1$ initial cells. Similarly, other cells may fire and
deplete the potential of the $n$-th cell. Again, assuming a homogenous and statistical identical population,
we have:
\begin{equation}
q=\frac{\lambda^++rq(n-1)\sum_{l=0}^\infty [\frac{qp(n-1)}{n}]^l\frac{1-p}{n}}{r+\lambda^-+q_uw^-_{u} + rq(n-1)\sum_{l=0}^\infty [\frac{qp(n-1)}{n}]^l\frac{p}{n}},
\end{equation}
becoming:
\begin{equation}
q=\frac{\lambda^++\frac{rq(n-1)(1-p)}{n-qp(n-1)}}{r+\lambda^-+q_uw^-_u+\frac{rqp(n-1)}{n-qp(n-1)}}, \label{q}
\end{equation}
which is a second degree polynomial in $q$:
\begin{eqnarray} \label{eqn.cluster1}
q^2p(n-1)[\lambda^-+q_uw^-_u]+q(n-1)[r(1-p)-\lambda^+p]\\
-q n (r+\lambda^- + q_uw^-_u) + \lambda^+ n =0. \nonumber
\end{eqnarray}
and its positive root which is less than one is computed, since the value of $q$ that we seek is a probability.
Let $x=q_uw^-_u$, then
\begin{equation}\label{eqn.activation}
\zeta(x)=\frac{-(C-nx)-\sqrt{(C-nx)^2-4 p(n-1)(\lambda^-+x)d}}{2 p(n-1)(\lambda^-+ x)},
\end{equation}
$d=n \lambda^+$ and
$C=\lambda^+ p + r p -\lambda^- n -r - \lambda^+ p n -n p r$.
For notation ease, we will use $\zeta(\cdot)$ as a term-by term-function for vectors and matrices.

\section{Improved Training Procedure for RNN-MLA} \label{sec.improve_train}

\begin{algorithm}[t]
\caption{Improved training procedure for the RNN-MLA}\label{algorithm2}
\begin{algorithmic}
\STATE Get data matrix $X$ and label matrix $Y$
\STATE \textbf{for} $l=1,\cdots,L-1$ \textbf{do}
\STATE ~~~~~solve Problem (\ref{eqn.problem3}) for $W_{l}$ with input $X$
\STATE ~~~~~$W_{l} \leftarrow W_{l}/\max(\zeta(X W_{l}))/10$
\STATE ~~~~~$X \leftarrow \zeta(X W_{l})$
\STATE randomly generate $W_{L}$ in range $[0~1]$
\STATE $W_{L+1} \leftarrow \text{pinv}(\zeta(X W_{L})) Y$
\end{algorithmic}
\end{algorithm}

In this section, we improve the training procedure for the RNN-MLA by modifying the reconstruction used in \cite{gelenbedeep2016}, normalizing RNN cluster inputs and adjusting external arrival rates of spikes $\lambda^+$ and $\lambda^-$ inside clusters.
Let us denote the connecting weight matrices between layers of the $L$-hidden-layer ($L \geq 2$) RNN-MLA by $W_1,\cdots,W_{L} \geq 0$ and output weight matrix by $W_{L+1}$. Adapted from \cite{gelenbedeep2016}, the weights $W_{l}$ $(l=1,\cdots,L-1)$ are determined by solving an reconstruction problem:
\begin{eqnarray} \label{eqn.problem3}
&& \min_{W_l} ||X -  \text{adj}(\zeta(X \bar{W})) W_{l} ||^2+||W_{l}||_{\ell_1}, ~ \text{s.t.~}W_{l} \geq 0,
\end{eqnarray}
where $\bar{W} \geq 0$ is randomly generated, operation adj$(X)$ first maps its input into $[0~1]$ linearly, then uses the ``zcore'' MATLAB operation and finally adds a positive constant to remove negativity. The fast iterative shrinkage-thresholding
algorithm (FISTA) in \cite{FISTA} is used to solve Problem (\ref{eqn.problem3}) for $W_{l}$ with the modification of setting negative elements in the solution to zero in each iteration. Weight matrix $W_{L}$ is randomly generated in range $[0~1]$, while $W_{L+1}$ is determined by the Moore-Penrose pseudo-inverse \cite{gelenbedeep2016,zhang2014cross,yin2012weights,zhang2012pruning,elm,mlelm} (denoted by ``pinv''). The improved training procedure for the RNN-MLA is shown in Algorithm \ref{algorithm2}, where operation $\max(\cdot)$ produces the maximal element of its input. Numerous numerical tests show that $0.01,0.005$ are generally good choices for $\lambda=\lambda^+=\lambda^-$.

\begin{figure}[t]
\centering
\includegraphics[width=4in]{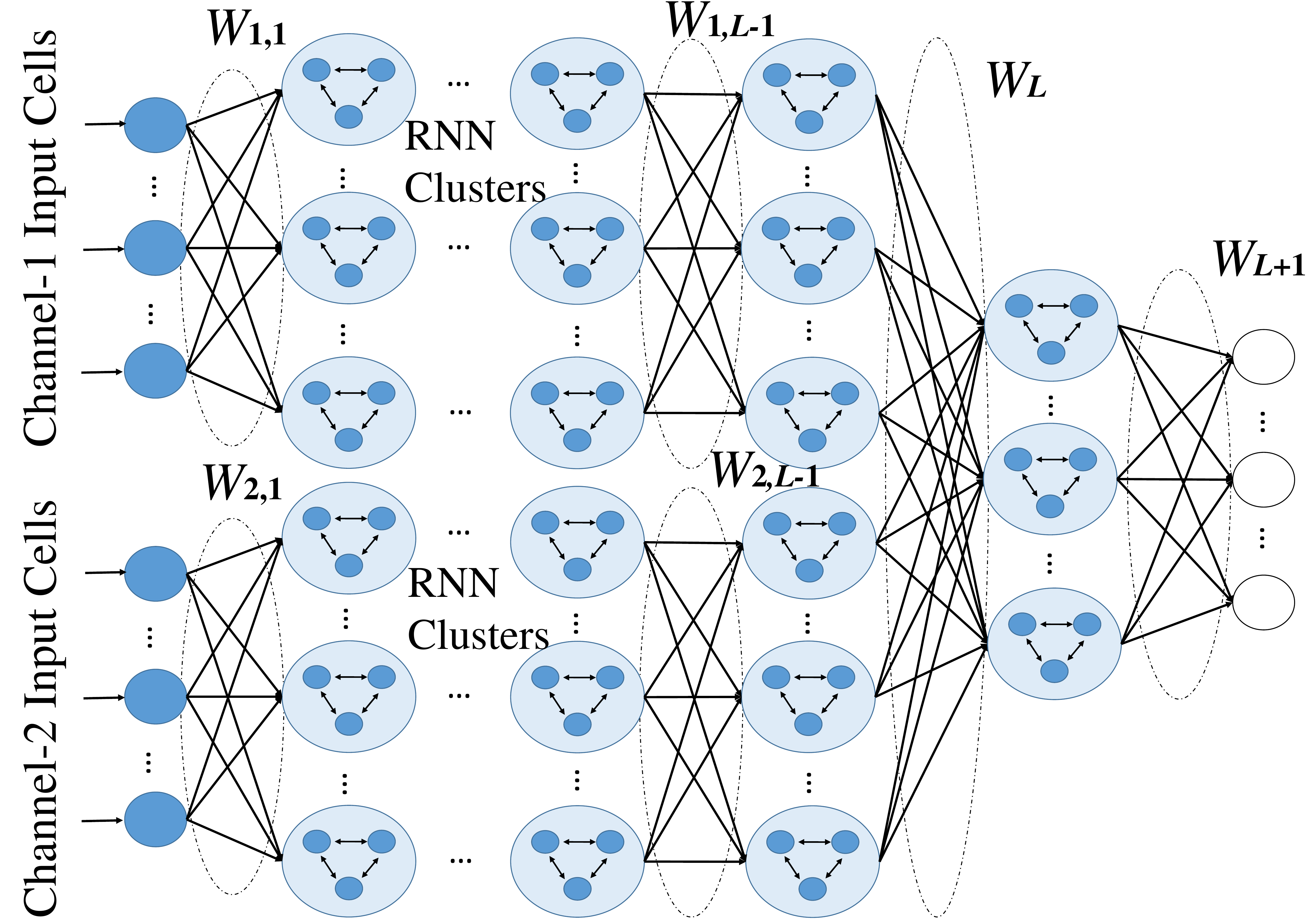}
\caption{Schematic representation of the MCRNN-MLA.}
\label{fig.MCRNN}
\end{figure}

\begin{algorithm}[t]
\caption{Training procedure for the MCRNN-MLA}\label{algorithm3}
\begin{algorithmic}
\STATE Get data matrices $X_c$ ($c=1,\cdots,C$) and label matrix $Y$
\STATE \textbf{for} $l=1,\cdots,L-1$ \textbf{do}
\STATE ~~~~~\textbf{for} $c=1,\cdots,C$ \textbf{do}
\STATE ~~~~~~~~~~solve Problem (\ref{eqn.problem4}) for $W_{c,l}$ with input $X_c$
\STATE ~~~~~~~~~~$W_{c,l} \leftarrow W_{c,l}/\max(\zeta(X_c W_{c,l}))/10$
\STATE ~~~~~~~~~~$X_c \leftarrow \zeta(X_c W_{c,l})$
\STATE $X \leftarrow [X_1 ~\cdots~X_C]$
\STATE randomly generate $W_{L}$ in range $[0~1]$
\STATE $W_{L+1} \leftarrow \text{pinv}(\zeta(X W_{L})) Y$
\end{algorithmic}
\end{algorithm}

\section{RNN-MLA for Multi-Channel Classification Datasets}

We now adapt the RNN-MLA to handle multi-channel classification datasets, called the MCRNN-MLA.
The superiority of the MCRNN-MLA is then demonstrated by results on both multi-channel image and real-world classification datasets.
For ease of illustration, we consider a dataset with two channels (Channel-1 and 2), a 2-channel $L$-hidden-layer MCRNN-MLA for which is shown schematically in Figure \ref{fig.MCRNN}.

Let us denote the connecting weights between layers for only Channel-1 by $W_{1,1},\cdots$, $W_{1,L-1} \geq 0$, those for only Channel-2 by $W_{2,1},\cdots,W_{2,L-1} \geq 0$, those between the $L-1$ and $L$ hidden layers by $W_{L}\geq 0$ and output weights by $W_{L+1}$.
The weights $W_{c,l} \geq 0$ ($c=1,2; l=1,\cdots,L-1$) are determined by solving an reconstruction problem using the modified FISTA (described in Section \ref{sec.improve_train}):
\begin{eqnarray} \label{eqn.problem4}
&& \min_{W_{c,l}} ||X_c -  \text{adj}(\zeta(X_c \bar{W})) W_{c,l} ||^2+||W_{c,l}||_{\ell_1}, ~ \text{s.t.~}W_{c,l} \geq 0,
\end{eqnarray}
where $X_c$ is either the data from Channel-$c$ or its layer encodings.
The training procedure of a $C$-channel $L$-hidden-layer MCRNN-MLA is shown in Algorithm \ref{algorithm3}.

\subsection{Modifications to the MCRNN-MLA}

\begin{figure}[t]
\centering
\includegraphics[width=4.5in]{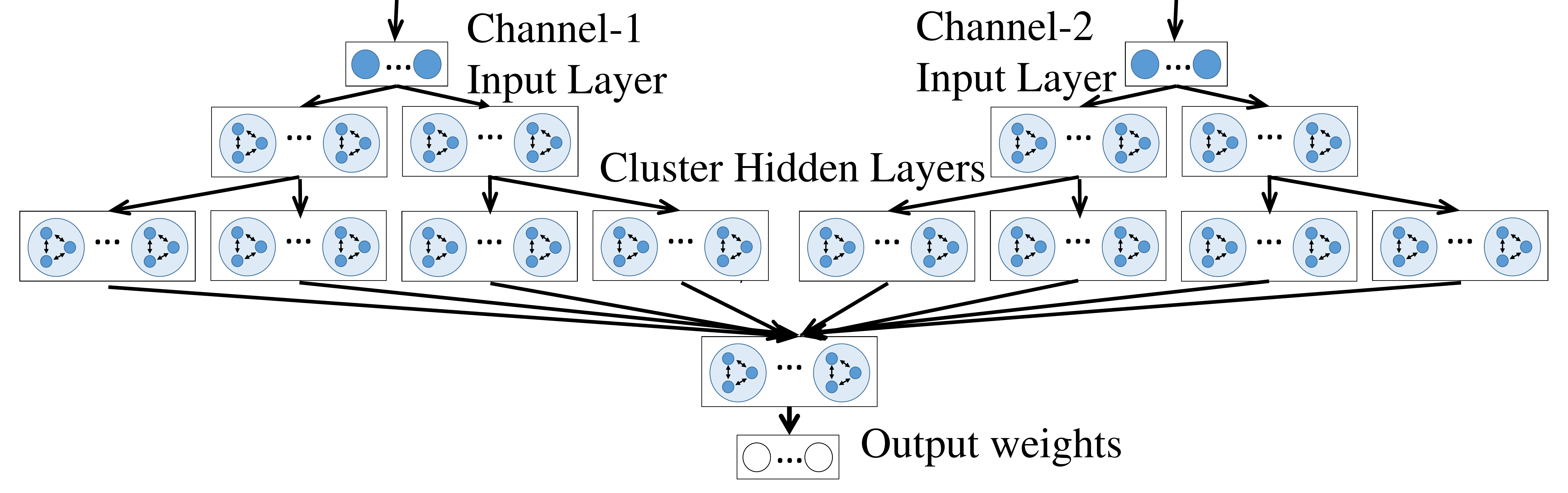}
\caption{Schematic representation of the MCRNN-MLA1.}
\label{fig.MCRNN2}
\end{figure}

We make a first modifications to the MCRNN-MLA, and call it MCRNN-MLA1, where the schematic representation of a $C$-channel $L$-hidden-layer $B$-branch MCRNN-MLA1 is shown in Figure \ref{fig.MCRNN2}.
Let us denote the connecting weights to the $l$th hidden layer for Channel-$c$ of Branch-$b$ by $W_{c,l,b} \geq 0$ ($c=1,\cdots,C; l=1,\cdots,L-1; b=1,\cdots,B$), those for all channels between the $L-1$ and $L$ hidden layers by $W_{L}\geq 0$ and output weights by $W_{L+1}$.
The training procedure of the MCRNN-MLA1 is detailed in Algorithm \ref{algorithm4}.

\begin{algorithm}[t]
\caption{Training procedure for the MCRNN-MLA1}\label{algorithm4}
\begin{algorithmic}
\STATE Get data matrices $X_c$ and let $X_{c,b} \leftarrow X_c$ for $b=1,\cdots,B$ ($c=1,\cdots,C$), and get label matrix $Y$;
\STATE \textbf{for} $l=1,\cdots,L-1$ \textbf{do}
\STATE ~~~~~\textbf{for} $c=1,\cdots,C$ \textbf{do}
\STATE ~~~~~~~~~~\textbf{for} $b=1,\cdots,B$ \textbf{do}
\STATE ~~~~~~~~~~~~~~~solve an similar reconstruction problem to (\ref{eqn.problem4}) for $W_{c,l,b}$ with input $X_{c,b}$
\STATE ~~~~~~~~~~~~~~~$W_{c,l,b} \leftarrow W_{c,l,b}/\max(\zeta(X_{c,b} W_{c,l,b}))/10$
\STATE ~~~~~~~~~~~~~~~$X_{c,b} \leftarrow \zeta(X_{c,b} W_{c,l,b})$
\STATE $X \leftarrow [X_{c,b}]$ for $b=1,\cdots,B$ and $c=1,\cdots,C$
\STATE randomly generate $W_{L}$ in range $[0~1]$
\STATE $W_{L+1} \leftarrow \text{pinv}(\zeta(X W_{L})) Y$
\end{algorithmic}
\end{algorithm}

\begin{table}[t]
\caption{Testing accuracies ($\%$) and training time (s) of different methods for NORB and DAS datasets.} \label{tab.norbsport}
\begin{center}
\begin{tabular}{|l|l|l|l|l|}
\hline
\multirow{2}*{Method}  &\multicolumn{2}{|l|}{Testing accuracy} &\multicolumn{2}{|l|}{Training time} \\\cline{2-5}
 &NORB &DAS&NORB &DAS\\	\hline
MCRNN-MLA&\textbf{92.10}&\textbf{99.21} &28.80&26.81\\	\hline
MCRNN-MLA1&91.21&98.98&1750.85&89.16\\	\hline
MCRNN-MLA2&91.72&94.67&1168.61&177.03\\	\hline
Improved RNN-MLA&90.96&92.17&20.63&13.11\\	\hline
Original RNN-MLA \cite{gelenbedeep2016}&88.51	&92.83&18.80&6.02\\	\hline
MLP$+$dropout \cite{chollet2015keras}&67.12 &91.94&2563.27&3291.47	\\	\hline
CNN \cite{chollet2015keras}&90.80 &98.52&1223.93&1289.76	\\	\hline
CNN$+$dropout \cite{chollet2015keras}&90.76 &99.05&1282.99&1338.35\\	\hline
H-ELM  \cite{mlelm}&87.56 &96.58&125.86&9.60\\	\hline
H-ELM *  \cite{mlelm}&91.28 &--&--&--\\	\hline
\multicolumn{3}{l}{*This data is obtained directly from \cite{mlelm}.}
\end{tabular}
\end{center}
\end{table}

The second modification denoted MCRNN-MLA2 is a simplified MCRNN-MLA1, obtained by removing the last hidden layerof MCRNN-MLA1 that produces random features via random connections $W_{L}$. The schematic representation and training procedure is omitted because it is similar to the previous one.

\subsection{Numerical Result Comparisons}\label{sec.MCRNNMLAresult}

We now move to numerical tests that use three multi-channel classification datasets: an image dataset and two real-world time-series datasets.

\subsubsection{NORB Dataset} The small NORB dataset \cite{lecun2004learning} is intended for experiments in 3D object recognition from shape. The instance numbers for both training and testing are 24300. There are two $96 \times 96 $ images in each instance which are downsampled into $32 \times 32$. All images are whitened using the code provided by \cite{mlelm}.

\subsubsection{Daily and Sports Activities (DSA) Dataset}
The DSA dataset \cite{altun2010comparative,barshan2014recognizing,altun2010human} comprises time-series data of 19 daily and sports activities performed by 8 subjects recorded by 45 motion sensors (25 Hz sampling frequency). The attribute number is 5,625 (45x5x25) since 5-second segments are used, while the class number is 19. Two thirds of 9120 instances are used for training while the rest for testing.

\subsubsection{Twin Gas Sensor Arrays (TGSA) Dataset}
The TGSA dataset includes 640 recordings of 5 twin 8-sensor detection units exposing to 4 different gases \cite{fonollosa2016calibration}.
The duration of each recording is 600 seconds (100Hz sampling frequency) producing 480,000 (8x600x100) features. We use 30-second segments, and then each instance has 24,000 (8x3000) attributes. The objective is to classify gas types using recording features. Two tasks are conducted, in both of which two thirds of instances are used for training while the rest for testing:
\begin{itemize}
\item Task 1: (3,029 instances): build a specific classifier for Unit 1 to fulfill the objective.
\item Task 2: (21,169 instances): build one classifier for all units to fulfill the objective.
\end{itemize}

The numbers of channels in the NORB, DSA and TGSA datasets are 2, 45 and 8, respectively. In the numerical experiments, we use the MCRNN-MLA, MCRNN-MLA1, MCRNN-MLA2, RNN-MLA with Algorithm \ref{algorithm2}, as well as the algorithm that was reported in \cite{gelenbedeep2016}, and the multi-layer perception (MLP)  from \cite{chollet2015keras}, the convolutional neural network (CNN) \cite{chollet2015keras,srivastava2014dropout} and hierarchical
extreme learning machine (H-ELM) \cite{mlelm}.

The results are summarised in Tables \ref{tab.norbsport} and \ref{tab.gas}. We can see that in most cases the improved RNN-MLA of this paper providesbetter results than the original one from \cite{gelenbedeep2016}. The proposed MCRNN-MLA (or its modification) achieves the highest testing accuracies for all cases. Moreover, the MCRNN-MLA can be trained much faster than the MLP and CNN. For example, it is trained around 127 times faster than the CNN for Task 2 of the TGSA dataset.
These results show that the MCRNN is the better tool for handling the classification of multi-channel datasets.

\begin{table}[t]
\caption{Testing accuracies ($\%$) and training time (s) of different methods for TGSA dataset.} \label{tab.gas}
\begin{center}
\begin{tabular}{|l|l|l|l|l|}
\hline
\multirow{2}*{Method}  &\multicolumn{2}{|l|}{Testing accuracy} &\multicolumn{2}{|l|}{Training time} \\\cline{2-5}
 &Task 1 &Task 2&Task 1 &Task 2\\	\hline
MCRNN-MLA&98.32&\textbf{92.25} &29.56&106.88\\	\hline
MCRNN-MLA1&\textbf{98.61}&90.11&55.08&143.76\\	\hline
MCRNN-MLA2&94.75&79.52&51.96&75.20\\	\hline
Improved RNN-MLA&97.03&87.06&16.78&149.47\\	\hline
Original RNN-MLA \cite{gelenbedeep2016}&85.64	&80.29&29.94&154.00\\	\hline
MLP$+$dropout \cite{chollet2015keras}&25.05 &24.86&3327.52&9005.39	\\	\hline
CNN \cite{chollet2015keras}&61.78 &72.13&1842.38&13593.06	\\	\hline
CNN$+$dropout \cite{chollet2015keras}&69.11 &87.00&2484.18&15545.18\\	\hline
H-ELM  \cite{mlelm}&61.98 &55.94&14.21&122.02\\	\hline
\end{tabular}
\end{center}
\end{table}

\section{Conclusions}

In previous work \cite{gelenbedeep2016} we had proposed the RNN-MLA and demonstrated its usefulness in deep learning with several data sets. In this paper, we have improved the training procedure for the RNN-MLA and proposed the novel MCRNN-MLA for classifying multi-channel datasets.

Comparative numerical experiments were conducted using several  different deep-learning methods based on both images and real-world data for multi-channel datasets. The numerical results show that the proposed MCRNN-MLA provides useful improvements as compared to other methods, in terms of both classification accuracy and training efficiency.


\bibliographystyle{IEEEtran}
\bibliography{RNN}
\end{document}